\documentclass[sigconf, screen, dvipsnames, authorversion]{acmart}

\usepackage[noend]{algorithmic}
\usepackage{algorithm}
\usepackage{multirow}
\usepackage{amsmath}
\usepackage{bbm}
\usepackage{booktabs}
\usepackage[inline]{enumitem} 
\usepackage{subcaption}
\usepackage{bm} 
\usepackage{xcolor}

\usepackage{hyperref}
\usepackage{lipsum}

\usepackage{wrapfig}

\usepackage{arydshln}
\makeatletter
\def\adl@drawiv#1#2#3{
        \hskip.5\tabcolsep
        \xleaders#3{#2.5\@tempdimb #1{1}#2.5\@tempdimb}%
                #2\z@ plus1fil minus1fil\relax
        \hskip.5\tabcolsep}
\newcommand{\cdashlinelr}[1]{%
  \noalign{\vskip\aboverulesep
          \global\let\@dashdrawstore\adl@draw
          \global\let\ adl@draw\adl@drawiv}
  \cdashline{#1}
  \noalign{\global\let\adl@draw\@dashdrawstore
          \vskip\belowrulesep}}
\makeatother

\usepackage{tikz}
\usetikzlibrary{automata, arrows, bayesnet, bending}

\usepackage{tikzscale}

\DeclareMathOperator*{\argmin}{arg\,min}

\copyrightyear{2024}
\acmYear{2024}
\setcopyright{acmlicensed}\acmConference[RecSys '24]{18th ACM Conference on Recommender Systems}{October 14--18, 2024}{Bari, Italy}
\acmBooktitle{18th ACM Conference on Recommender Systems (RecSys '24), October 14--18, 2024, Bari, Italy}
\acmDOI{10.1145/3640457.3688162}
\acmISBN{979-8-4007-0505-2/24/10}
\begin{document}

\title{$\Delta{\rm\text{-}OPE}$: Off-Policy Estimation with Pairs of Policies}

\author{Olivier Jeunen}
\affiliation{
  \institution{ShareChat}
  \country{Edinburgh, United Kingdom}
}  \email{jeunen@sharechat.co}

\author{Aleksei Ustimenko}
\affiliation{
  \institution{ShareChat}
  \country{London, United Kingdom}
}  \email{aleksei.ustimenko@sharechat.co}

\begin{abstract}
The off-policy paradigm casts recommendation as a counterfactual decision-making task, allowing practitioners to unbiasedly estimate online metrics using offline data.
This leads to effective evaluation metrics, as well as learning procedures that directly optimise online success.
Nevertheless, the high variance that comes with unbiasedness is typically the crux that complicates practical applications.
An important insight is that the \emph{difference} between policy values can often be estimated with significantly reduced variance, if said policies have positive covariance. 
This allows us to formulate a pairwise off-policy estimation task: $\Delta\text{-}{\rm OPE}$.

$\Delta\text{-}{\rm OPE}$ subsumes the common use-case of estimating improvements of a \emph{learnt} policy over a \emph{production} policy, using data collected by a stochastic \emph{logging} policy.
We introduce $\Delta\text{-}{\rm OPE}$ methods based on the widely used Inverse Propensity Scoring estimator and its extensions.
Moreover, we characterise a variance-optimal additive control variate that further enhances efficiency.
Simulated, offline, and online experiments show that our methods significantly improve performance for both evaluation and learning tasks.
\end{abstract}

\maketitle

\section{Introduction \& Motivation}
Recommendation tasks are increasingly often modelled through a decision-making lens~\cite{Vasile2021_DecisionTheory}, leveraging ideas from causal and counterfactual inference to improve \emph{evaluation} and \emph{learning} procedures prevalent in the field~\cite{Joachims2018_REVEAL,Vasile2020,Saito2021,CONSEQUENCES2022,Gupta2024}.
The \emph{off-policy estimation} (OPE) paradigm is especially attractive to practitioners~\cite{vandenAkker2024}, as it allows key online metrics to be estimated from offline data alone~\cite{Jeunen2021_Thesis}.
This has sparked significant research interest, as well as practical advances to successfully apply such methods to broad recommendation use-cases~\cite{Bottou2013, Yang2018,Gilotte2018, Gruson2019, chen2019top, Dong2020, ma2020off, Jeunen2020, Jeunen2021_Pessimism, Chen2021, Mairesse2021, Liu2022, chen2022actorcritic, Jeunen2023,gupta2023safe,gupta2023deep}.

The well-known bias-variance trade-off is at the heart of what complicates practical deployment of such methods: unbiased estimation methods (based on importance sampling~\cite{Horvitz1952}) suffer from high variance, giving theoretical guarantees but confidence intervals that are too wide to be useful for practical decision-making.
Methods that reduce estimation variance are thus key to success.

Our work leverages the observation that whilst the counterfactual value-of-interest is typically framed as: ``\emph{What would the value for the online metric have been, had we deployed the target policy instead?}''; the decision-making criterion for shipment in industrial applications is rather: ``\emph{What is the \textbf{difference} between the online metric value under the target policy, and the production policy?}''

Whilst this discrepancy can seem subtle, it has implications for the efficiency of our statistical estimators: if the production and target policies are correlated, the latter estimation task can be performed with lower variance and, hence, tighter confidence intervals.
This leads to improved statistical power (reduced type-II error) in \emph{evaluation} scenarios~\cite{Gilotte2018}, and improved recommendation policies in counterfactual learning scenarios~\cite{Swaminathan2015_BLBF}.
Our work introduces this $\Delta$-OPE task, extending existing OPE methods and deriving the optimal unbiased estimator with a global additive control variate.
Empirical insights from reproducible simulations and large-scale online A/B-experiments align with our theoretical expectations.
\section{Methodology \& Contributions}
We deal with a general contextual bandit setup, as commonly applied to recommendation scenarios.
The context is described by a random variable $X$, typically describing historical and current user features.
Actions $A$ are potential items being recommended, and the recommender system itself is defined as a policy that describes a conditional probability distribution over actions given context: $\mathsf{P}(A=a|X=x;\Pi=\pi)\equiv\pi(a|x)$.
Recommendations lead to rewards $R$, which can comprise several types of interactions in the most general sense (i.e. clicks, purchases, likes, streams, et cetera).

The value of a policy is defined as the reward it yields.
With an expectation over contexts $x \sim \mathsf{P}(X)$, actions $a \sim \pi(\cdot|x)$, and rewards $r \sim \mathsf{P}(R|X=x;A=a)$, we write $V(\pi) = \mathop{\mathbb{E}}[r]$.

OPE methods leverage data collected by a stochastic logging policy $\pi_{0}$ to estimate the value of some target policy $\pi_{t}$.
Inverse Propensity Score (IPS) weighting is the bread and butter that enables this family of methods~\cite[\S 9]{Owen2013}.
With a dataset $\mathcal{D} \coloneqq \{(x_{i},a_{i},{r_{i}})_{i=1}^{N}\}$ collected under $\pi_{0}$, the unbiased IPS estimator is given by:
\begin{equation}
    \widehat{V}_{\rm IPS}(\pi_{t},\mathcal{D}) = \frac{1}{|\mathcal{D}|}\sum_{(x,a,r) 
    \in \mathcal{D}} \frac{\pi_{t}(a|x)}{\pi_{0}(a|x)}r.
\end{equation}
Whilst unbiased, $\widehat{V}_{\rm IPS}$ can suffer from high variance.
Multiplicative control variates ---a common tool for variance reduction--- give rise to the widely used Self-Normalised IPS (SNIPS) estimator~\cite[\S 9.2]{Owen2013}:
\begin{equation}
    \widehat{V}_{\rm SNIPS}(\pi_{t},\mathcal{D}) = \frac{\sum_{(x,a,r) 
    \in \mathcal{D}} \frac{\pi_{t}(a|x)}{\pi_{0}(a|x)}r}{\sum_{(x,a,r) 
    \in \mathcal{D}} \frac{\pi_{t}(a|x)}{\pi_{0}(a|x)}}.
\end{equation}
\citet{Gupta2024_OptimalBaselines} show that additive control variates are either competitive or superior, with the $\beta$-IPS estimator defined as:
\begin{gather}
    \widehat{V}_{\beta\text{-}{\rm IPS}}(\pi_{t},\mathcal{D}) = \beta + \frac{1}{|\mathcal{D}|}\sum_{(x,a,r) 
    \in \mathcal{D}} \frac{\pi_{t}(a|x)}{\pi_{0}(a|x)}(r-\beta),\label{eq:beta_IPS}\\
    \text{with~} \beta =  \frac{\sum_{(x,a,r) \in \mathcal{D}} \left(\left(\frac{ \pi_{t}(a|x)}{\pi_{0}(a|x)}\right) ^2  - \frac{ \pi_{t}(a|x)}{\pi_{0}(a|x)}\right) r }{\sum_{(x,a,r) \in \mathcal{D}} \left(\frac{ \pi_{t}(a|x)}{\pi_{0}(a|x)}\right) ^2 - \left(\frac{ \pi_{t}(a|x)}{\pi_{0}(a|x)}\right)}.\label{eq:optimal_baseline}
\end{gather}
Indeed, the additive nature of the control variate preserves unbiasedness, and the value given by Eq.~\ref{eq:optimal_baseline} is an empirical estimate of the variance-minimising baseline correction.

Doubly robust methods add expressive power to the control variate by replacing it with a learnt reward model~\cite{Dudik2014}, and specialised methods to optimise said model have been proposed as well~\cite{Farajtabar2018}.
3Nevertheless, the optimisation of a reward model can be costly, and empirical improvements are not guaranteed~\cite{Jeunen2020REVEAL}.

In off-policy \emph{learning} scenarios, the Counterfactual Risk Minimisation (CRM) principle is often used to optimise a pessimistic lower bound on an off-policy estimator~\cite{Swaminathan2015,Swaminathan2015_BLBF,Jeunen2021_Pessimism,Jeunen2023,Jeunen2024_MultiObjective}, which is typically obtained through sample variance penalisation~\cite{Maurer2009}.

\subsection{Pairwise Off-Policy Estimation}
OPE methods typically focus on counterfactual estimation of a single policy value $V(\pi_{t})$, based on data collected under a stochastic logging policy $\pi_0$.
It is important to note, however, that the logging policy $\pi_{0}$ is seldom very competitive.
Indeed, there is typically a distinct \emph{production} policy $\pi_{p}$ in place; and $\pi_{0}$ adds increased stochasticity on top of it to allow for effective IPS-weighted estimation~\cite{Jeunen2020,Mairesse2021,Jeunen2024_MultiObjective, wan2022safe}.
Several works have even reported leveraging uniformly distributed logging policies~\cite{Li2010, Gruson2019, Dong2020, Liu2020, Sagtani2024}.

As such, the quantity we wish to estimate is not merely $V(\pi_{t})$, but the improvement $\pi_{t}$ might bring over the production policy $\pi_{p}$:
\begin{equation}
    V_{\Delta}(\pi_{t}, \pi_{p}) = V(\pi_{t}) - V(\pi_{p}) =\mathbb{E}_{a\sim \pi_{t}}[r] -\mathbb{E}_{a\sim \pi_{p}}[r].
\end{equation}
In the off-policy setting, we do not have access to samples from $\pi_{t}$.
Because $\pi_{p}$ is deployed in production, we do have access to samples from $\pi_{p}$.
Nevertheless, they might not be informative for counterfactual estimation of $V(\pi_{t})$ due to limited randomisation.
For this reason, we have an additional logging policy $\pi_{0}$.
The key insight is that we can also use data from the logging policy to estimate $V(\pi_{p})$.
This re-framing implies that the $V_{\Delta}(\pi_{t}, \pi_{p})$ estimand can be written as a single expectation over $\pi_{0}$, as:
\begin{gather}
V_{\Delta}(\pi_{t},\pi_{p}) = \mathop{\mathbb{E}}\limits_{a\sim \pi_{0}}\left[\frac{\pi_{t}(a|x)}{\pi_{0}(a|x)}r \right] - \mathop{\mathbb{E}}\limits_{a\sim \pi_{0}}\left[\frac{\pi_{p}(a|x)}{\pi_{0}(a|x)}r\right] = \nonumber\\
\mathop{\mathbb{E}}\limits_{a\sim \pi_{0}}\left[\frac{\pi_{t}(a|x)}{\pi_{0}(a|x)}r - \frac{\pi_{p}(a|x)}{\pi_{0}(a|x)}r\right] = \mathop{\mathbb{E}}\limits_{a\sim \pi_{0}}\left[\frac{(\pi_{t}(a|x) - \pi_{p}(a|x))}{\pi_{0}(a|x)}r\right].    
\end{gather}
As such, if the traditional IPS assumptions of common support and unconfoundedness hold~\cite{Owen2013, JeunenLondon2023} , we can derive unbiased estimators for this quantity.
The advantages of the $\Delta\text{-}{\rm OPE}$ framing become clear when we consider a decomposition of its variance:
\begin{gather}
{\rm Var}\left(V_{\Delta}(\pi_{t},\pi_{p})\right) = \nonumber\\ {\rm Var}\left(V(\pi_{t})\right)+{\rm Var}\left(V(\pi_{p})\right)-2\cdot{\rm Covar}\left(V(\pi_{t});V(\pi_{p})\right).    
\end{gather}
This gives rise to the following condition for variance reduction:
\begin{gather}{\rm Var}\left(V_{\Delta}(\pi_{t},\pi_{p})\right)  <\rm{Var}(V(\pi_{t})) \\\iff\nonumber\\{\rm Var}\left(V(\pi_{p})\right)<2\cdot{\rm Covar}\left(V(\pi_{t}),V(\pi_{p})\right).\label{eq:variance_reduction_condition}\end{gather}

If the inequality in Eq.~\ref{eq:variance_reduction_condition} holds, we can estimate $V_{\Delta}(\pi_{t},\pi_{p})$ with tighter confidence intervals than we would be able to estimate $V(\pi_{t})$.
This implies a more sample-efficient off-policy evaluation method compared to directly estimating $V(\pi_{t})$.
For learning scenarios that follow the CRM principle, $V_{\Delta}(\pi_{t},\pi_{p})$ will favour policies $\pi_{t}$ that have high covariance with the target policy $\pi_{p}$.
This implies strong theoretical connections to existing policy learning methods, e.g. based on distributionally robust~\cite{Faury2020,Si2020}, trust region~\cite{Schulman2015} or proximal optimisation~\cite{Schulman2017}.
Indeed: optimising a lower bound on $V_{\Delta}$ leads to a learnt target policy $\pi_{t}$ with probabilistically guaranteed improvements over the production policy $\pi_{p}$.
In what follows, we derive finite sample estimators for the $\Delta\text{-}{\rm OPE}$ task: first for classical IPS, then for multiplicative control variates with SNIPS, and additive control variates with $\beta$-IPS.
We characterise the optimal variance-minimising value of $\beta$ for the latter family of estimators.

\subsubsection{Pairwise Inverse Propensity Scoring: $\Delta{\rm\text{-}IPS}$}
The difference between IPS estimates can be written as:
\begin{gather}\widehat{V}_{\Delta{\rm-IPS}}(\pi_{t},\pi_{p},\mathcal{D})= \widehat{V}_{\rm IPS}(\pi_{t},\mathcal{D}) - \widehat{V}_{\rm IPS}(\pi_{p},\mathcal{D})\nonumber\\
= \frac{1}{|\mathcal{D}|} \sum_{(x,a,r) \in \mathcal{D}} \frac{\pi_{t}(a|x)-\pi_{p}(a|x)}{\pi_{0}(a|x)}r.\end{gather}
The sample variance for $\widehat{V}_{\Delta{\rm-IPS}}$ is given by:
\begin{gather}\widehat{{\rm Var}}\left(\widehat{V}_{\Delta{\rm-IPS}}(\pi_{t},\pi_{p},\mathcal{D})\right)= \nonumber\\\frac{1}{|\mathcal{D}|-1} \sum_{(x,a,r) \in \mathcal{D}}\left(\frac{\pi_{t}(a|x)-\pi_{p}(a|x)}{\pi_{0}(a|x)}r - \widehat{V}_{\Delta{\rm-IPS}}(\pi_{t},\pi_{p},\mathcal{D})\right)^{2}.\end{gather}
These expressions allow us to provide confidence intervals evaluating improvements over the production policy for any target policy that has common support with the logging policy (i.e. an off-policy \emph{evaluation} task), or to formulate a CRM lower bound and learn a policy that directly optimises improvements over production (i.e. an off-policy \emph{learning} task).

\textit{Equivalence to baseline corrections.}
An interesting equivalence arises when we consider the special case where $\pi_{p}\equiv\pi_{0}$.
Indeed, in this case, the $\Delta$-OPE task reduces to traditional OPE:
\begin{gather}
    V_{\Delta{\rm-IPS}}(\pi_{t},\pi_{0}) 
    = \mathop{\mathbb{E}}\limits_{a\sim\pi_{0}}\left[\frac{\pi_t}{\pi_0}R\right]-\mathop{\mathbb{E}}\limits_{a\sim\pi_{0}}\left[R\right].
\end{gather}
We observe that the key estimand remains the same, but a simple baseline correction is added on top.
This baseline is the empirical average of observed rewards under the logging policy---exactly what \citet{Williams1988} originally proposed to use for general on-policy reinforcement learning scenarios.
Subsequent works have proposed improvements~\cite{Dayan1991,Greensmith2004}, with \citet{Gupta2024_OptimalBaselines} recently characterising optimal values for the general OPE setting we consider in this work.

\subsubsection{Multiplicative Control Variates: $\Delta{\rm\text{-}SNIPS}$}
Now, we consider the self-normalised IPS estimator for the mean:
\begin{gather}\widehat{V}_{\Delta{\rm-SNIPS}}(\pi_{t},\pi_{p},\mathcal{D})= \widehat{V}_{\rm SNIPS}(\pi_{t},\mathcal{D}) - \widehat{V}_{\rm SNIPS}(\pi_{p},\mathcal{D})\nonumber\\ = \left( \frac{\sum_{(x,a,r) \in \mathcal{D}} \frac{\pi_{t}(a|x)}{\pi_{0}(a|x)}r}{\sum_{(x,a,r) \in \mathcal{D}} \frac{\pi_{t}(a|x)}{\pi_{0}(a|x)}}\right) - \left(\frac{\sum_{(x,a,r) \in \mathcal{D}} \frac{\pi_{p}(a|x)}{\pi_{0}(a|x)}r}{\sum_{(x,a,r) \in \mathcal{D}} \frac{\pi_{p}(a|x)}{\pi_{0}(a|x)}}\right).\end{gather}

The estimator for its variance, however, is less straightforward to compute.
Indeed, because it is a difference in ratio metrics, we need to resort to the Delta method to obtain an approximate estimator for its variance~\cite[\S 11]{Lehmann2005}.
For legibility, write the empirical SNIPS control variate as:
\begin{equation}    
\bar{w}(\pi,\mathcal{D}) = \frac{1}{|\mathcal{D}|} \sum_{(x,a,r) \in \mathcal{D}} \frac{\pi(a|x)}{\pi_{0}(a|x)}.
\end{equation}
Then, we can derive an estimator for the $\Delta$-SNIPS variance as:
\begin{gather}
g = \left[\frac{1}{\bar{w}(\pi_{t},\mathcal{D})}, - \frac{\widehat{V}_{\rm IPS}(\pi_{t},\mathcal{D})}{\bar{w}(\pi_{t},\mathcal{D})^{2}},-\frac{1}{\bar{w}(\pi_{p},\mathcal{D})},  \frac{\widehat{V}_{\rm IPS}(\pi_{p},\mathcal{D})}{\bar{w}(\pi_{p},\mathcal{D})^{2}}\right],\nonumber\\
\Sigma_{\Delta{\rm-SNIPS}}  = {\rm Covar}\left(\widehat{V}_{\rm IPS}(\pi_{t},\mathcal{D}); \bar{w}(\pi_{t},\mathcal{D}); \widehat{V}_{\rm IPS}(\pi_{p},\mathcal{D}); \bar{w}(\pi_{p},\mathcal{D})\right), \nonumber\\    
\widehat{\rm Var}\left(\widehat{V}_{\Delta{\rm-SNIPS}}(\pi_{t}, \pi_{p}, \mathcal{D})\right) = g\Sigma_{\Delta{\rm-SNIPS}} g^{\top}.
\end{gather}
Note that, whilst we can expect reduced variance compared to $\Delta$-IPS, the $\Delta$-SNIPS estimator suffers from the same flaws as the traditional SNIPS estimator: it is only \emph{asymptotically} unbiased~\cite{Kong1992}, and it inhibits mini-batch-based optimisation~\cite{Joachims2018}.

\subsubsection{Additive Control Variates: $\Delta\beta{\rm\text{-}IPS}$}
Recently, \citet{Gupta2024_OptimalBaselines} have shown that additive control variates are equally competitive, whilst having the additional advantage that closed-form analytical expressions for the \emph{optimal} (variance-minimising) control variate can be derived.
We can thus derive a simple estimator for the difference in policy values that retains IPS’ unbiasedness:
\begin{gather}\widehat{V}_{\Delta\beta{\rm-IPS}}(\pi_{t},\pi_{p},\mathcal{D}) = \frac{1}{|\mathcal{D}|} \sum_{(x,a,r) \in \mathcal{D}} \frac{\pi_{t}(a|x)-\pi_{p}(a|x)}{\pi_{0}(a|x)}(r-\beta).\end{gather}
Its unbiasedness is easily verified by linearity of expectation:
\begin{gather}
    \mathop{\mathbb{E}}\limits_{a\sim\pi_{0}}\left[\left(\frac{\pi_{t}(a|x) - \pi_{p}(a|x)}{\pi_{0}(a|x)}\right)\left(r-\beta\right)\right] = \overbrace{\mathop{\mathbb{E}}\limits_{a\sim\pi_{0}}\left[\frac{\pi_{t}(a|x)}{\pi_{0}(a|x)}r\right]}^{V(\pi_{t})} - \nonumber\\ \underbrace{\mathop{\mathbb{E}}\limits_{a\sim\pi_{0}}\left[\frac{\pi_{p}(a|x)}{\pi_{0}(a|x)}r\right]}_{V(\pi_{p})} -
\beta\underbrace{\mathop{\mathbb{E}}\limits_{a\sim\pi_{0}}\left[\frac{\pi_{t}(a|x)}{\pi_{0}(a|x)}\right]}_{\equiv 1} + 
\beta\underbrace{\mathop{\mathbb{E}}\limits_{a\sim\pi_{0}}\left[\frac{\pi_{p}(a|x)}{\pi_{0}(a|x)}\right]}_{\equiv 1}.
\end{gather}
We can write the variance of this estimator as:
\begin{gather}{\rm Var}\left(\widehat{V}_{\Delta\beta{\rm-IPS}}(\pi_{t},\pi_{p})\right) =\\\mathbb{E}\left[\left(\frac{\pi_t(a|x)-\pi_p(a|x)}{\pi_0(a|x)}(r-\beta)\right)^{2}\right]-\underbrace{\mathbb{E}\left[\frac{\pi_t(a|x)-\pi_p(a|x)}{\pi_0(a|x)}(r-\beta)\right]^{2}}_{V_{\Delta}(\pi_{t},\pi_{p})^{2}}.\nonumber\end{gather}
Considering minimisation of variance as a function of $\beta$, we have:
\begin{gather}
\argmin_{\beta} {\rm Var}\left(\widehat{V}_{\Delta\beta{\rm-IPS}}(\pi_{t},\pi_{p})\right) = \nonumber \\
\argmin_{\beta} \mathbb{E}\left[\left(\frac{\pi_t(a|x)-\pi_p(a|x)}{\pi_0(a|x)}\right)^{2}(r-\beta)^{2}\right].
\end{gather}
The partial derivative of this quantity with respect to $\beta$ is given by:
\begin{gather}
\frac{\partial \left(\mathrm{Var} \left( \widehat{V}_{\Delta\beta\text{-}{\rm IPS}}(\pi_{t},\pi_{p}) \right)\right)}{\partial \beta}  = 2 \mathop{\mathbb{E}}\left[  \left(\frac{\pi_{t}(a|x)-\pi_{p}(a|x)}{\pi_{0}(a|x)}\right) ^2\left( \beta -  r  \right)  \ \right].
\end{gather}
Solving for the derivative to equal 0, yields the optimal baseline:
\begin{gather}
\beta^{*} = \frac{\mathop{\mathbb{E}}\limits_{a\sim\pi_{0}}\left[  \left(\frac{\pi_{t}(a|x)-\pi_{p}(a|x)}{\pi_{0}(a|x)}\right)^{2} r  \right]}{\mathop{\mathbb{E}}\limits_{a\sim\pi_{0}}\left[ \left(\frac{\pi_{t}(a|x)-\pi_{p}(a|x)}{\pi_{0}(a|x)}\right)^{2}  \right]}.
\end{gather}
An asymptotically unbiased estimate for the optimal value can be obtained from empirical samples under the logging policy.
This characterises the variance-optimal additive control variate for the $\Delta\beta$-IPS estimator.
As the estimator is unbiased, variance-optimality implies estimation-error-optimality in terms of mean squared error.
\begin{figure*}[t]
    \centering
    \begin{subfigure}[t]{\linewidth}
        \centering
        \vspace{-3ex}
        \includegraphics[width=\linewidth]{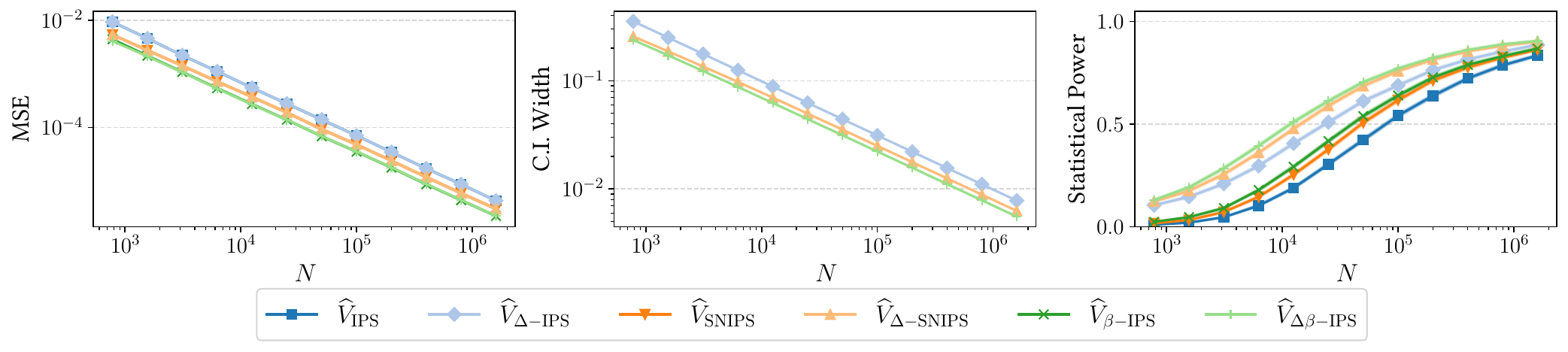}
        \caption{Off-policy evaluation results using the Open Bandit Pipeline to simulate pairs of recommendation policies.}
        \label{fig:OBP_discrete}
        \end{subfigure}%
~\\
    \begin{subfigure}[t]{\linewidth}
    \centering
    \includegraphics[width=\linewidth]{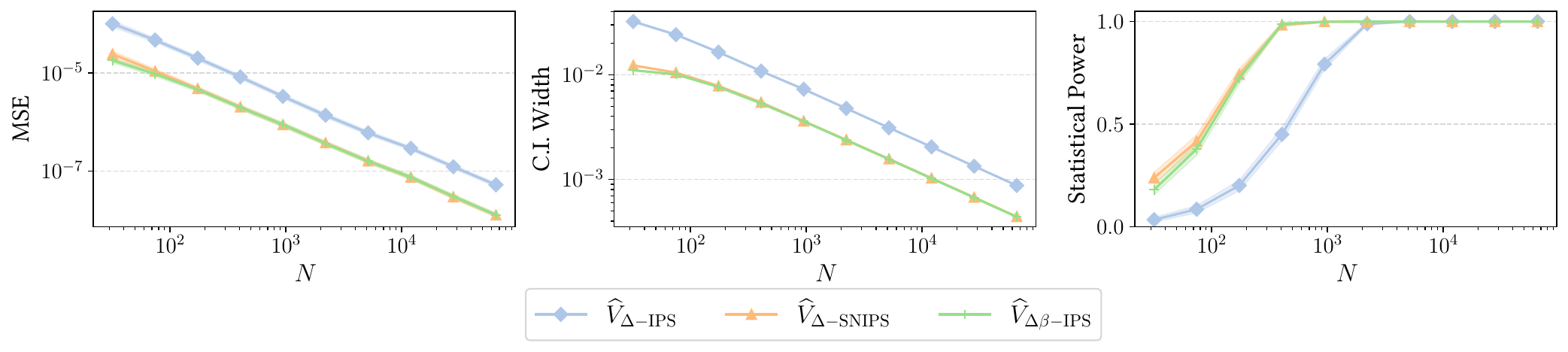}
    \caption{Off-policy evaluation results using synthetic continuous data to simulate pairs of recommendation policies.}
    \label{fig:continuous}
    \end{subfigure}%
    \caption{The $\Delta\text{-}{\rm OPE}$ estimator family significantly improves performance, with $\Delta\beta\text{-}{\rm IPS}$ consistently performing  best.}
\end{figure*}
\begin{table*}[!t]
\begin{subtable}[t]{0.45\textwidth}
    \centering
    \begin{tabular}{lc}
    \toprule
        \textbf{Metric} & \textbf{Relative Improvement (95\% C.I.)} \\
    \midrule
    Retention & \colorbox{gray!10}{[--0.05\%, +0.07\%]} \\
    Engaging Users & \colorbox{gray!10}{[--0.12\%, +0.12\%]} \\
    Learnt Reward & \colorbox{LimeGreen!50}{[+0.04\%, +0.27\%]}  \\
    \bottomrule
    \end{tabular}
    \caption{Optimised for $\widehat{V}_{\rm SNIPS}$ lower bound, 14 days, 6.4 million users.}
    \label{tab:AB1}
\end{subtable}
\hspace{0.05\textwidth}
~\hfill~
\begin{subtable}[t]{0.48\textwidth}
    \centering
    \begin{tabular}{lc}
    \toprule
        \textbf{Metric} & \textbf{Relative Improvement (95\% C.I.)} \\
    \midrule
    Retention & \colorbox{gray!10}{[--0.04\%, +0.06\%]} \\
    Engaging Users & \colorbox{LimeGreen!50}{[+0.00\%, +0.19\%]} \\
    Learnt Reward & \colorbox{LimeGreen!50}{[+0.14\%, +0.32\%]}  \\
    \bottomrule
    \end{tabular}
    \caption{Optimised for $\widehat{V}_{\Delta{\rm\text{-}SNIPS}}$ lower bound, 14 days, 14.8 million users.}
    \label{tab:AB2}
\end{subtable}
\caption{Online A/B-test results from deploying our approach on a large-scale short-video platform.
We report Bonferroni-corrected 95\% confidence intervals for key platform metrics.
We observe statistically significant improvements to both the optimisation target and adjacent metrics, \colorbox{LimeGreen!50}{highlighting} results that are statistically significant with $p<0.05$.}\label{tab:online_experiment}
\end{table*}
\section{Experiments \& Discussion}\label{sec:experiments}
We wish to answer the following research questions empirically:
\begin{description}
    \item[\textbf{RQ1}] \textit{Do our proposed $\Delta$-OPE methods improve statistical power in off-policy evaluation settings with discrete action spaces?}
    \item[\textbf{RQ2}] \textit{Do our proposed $\Delta$-OPE methods improve statistical power in off-policy evaluation settings with continuous action spaces?}
    \item[\textbf{RQ3}] \textit{Do our proposed $\Delta$-OPE methods lead to improved learnt recommendation policies in off-policy learning settings?}
\end{description}
The literature applying off-policy estimation methods to recommendation tasks typically uses either simulation methods, or online experiments.
Simulations have the advantage of reproducibility and controllability, whilst being potentially biased due to disparities between simulation assumptions and real-world users.
Online experiments have the advantage of directly measuring the quantities we care about, whilst being inherently irreproducible and inaccessible for most researchers.
In this work, we leverage both.

We provide all implementation details for the simulated experiments, and all source to reproduce the results on synthetic data can be found at \href{https://github.com/olivierjeunen/delta-OPE-recsys-2024}{github.com/olivierjeunen/delta-OPE-recsys-2024}.

\subsection{Evaluation with discrete actions (RQ1)}
We use the Open Bandit Pipeline to run reproducible simulation experiments that adhere to realistic recommendation use-cases~\cite{Saito2021_OBP}.
We vary the size of the action space $|\mathcal{A}| \in \{5, 10, 15\}$, the inverse temperature of a softmax logging policy as $\{1, 2, 4\}$, and the number of test set samples in $[800, 1\,600\,000]$.
We simulate a synthetic training dataset, which we use to learn target policies that maximise the $\widehat{V}_{\rm IPS}$ estimator.
The target policies use two different underlying models: logistic regression $\pi_{\rm lr}$, and a random forest $\pi_{\rm rf}$.
The advantage of the simulation environment is that we can obtain the true value difference $V_{\Delta}(\pi_{\rm lr}, \pi_{\rm rf})$, and use it as ground truth for the estimators we wish to evaluate.
We evaluate pointwise OPE methods: IPS, SNIPS and $\beta$-IPS; as well as our proposed pairwise alternatives: $\Delta$-IPS, $\Delta$-SNIPS and $\Delta\beta$-IPS.
We evaluate the estimators' Mean Squared Error (MSE), the width of the obtained confidence intervals, and the statistical power (i.e. $1-$Type-II error).
We should expect MSE to match between pointwise methods and their pairwise alternatives, as the mean of the estimator is simply the difference of the pairwise estimators.
Because pointwise methods do not estimate pairwise differences, we cannot visualise the width of the confidence interval for them.
Nevertheless, if the estimates for $\widehat{V}(\pi_{\rm lr})$ and $\widehat{V}(\pi_{\rm rf})$ have non-overlapping confidence intervals, we can claim a statistically significant difference.
For pairwise methods, this occurs when  $\widehat{V}_{\Delta}(\pi_{\rm lr},\pi_{\rm rf})$ does not include 0.
We repeat this procedure $1\,000$ times to smooth out randomisation noise, and report 95\% confidence intervals for relevant metrics in Figure~\ref{fig:OBP_discrete}.
We observe that the $\Delta\text{-}{\rm OPE}$ estimator family significantly improves performance, with $\Delta\beta\text{-}{\rm IPS}$ exhibiting the lowest MSE, with the tightest confidence intervals and highest statistical power; corroborating theoretical results.

\subsection{Evaluation with continuous actions (RQ2)}
Continuous action spaces also arise in recommendation use-cases, notably when optimising scalarisation weights in multi-objective recommendation scenarios~\cite{Jeunen2024_MultiObjective}.
We set up simple reproducible simulation experiments that emulate this setting, implemented in Python3.9 and leveraging the Numpy and Scipy libraries.
The logging policy $\pi_{0}$ is a $5$-dimensional isotropic Gaussian distribution centred at the constant $0.475$ vector, with covariance $0.5\cdot I$.
The production and target policies have shifted means at $0.5$ and $0.505$ respectively, and the reward function is simply the mean of the action vector values with added Gaussian noise.
As the reward is the mean of the vectors, the expected reward difference under the two policies is $V_{\Delta}(\pi_{t},\pi_{p}) = 0.005$.
Observed rewards have additive Gaussian noise, sampled from $\mathcal{N}(0,0.25)$.
As for RQ1, we repeat the procedure of sampling a dataset for off-policy estimation and evaluation of our estimators $1\,000$ times to smooth out randomisation noise, and report 95\% confidence intervals for relevant metrics in Figure~\ref{fig:continuous}.

Figure~\ref{fig:continuous} visualises these results, pointwise OPE methods are not visualised as their statistical power was 0 (i.e. all considered settings led to overlapping confidence intervals).
We observe that the $\Delta\text{-}{\rm OPE}$ estimator family significantly improves performance, with $\Delta$-SNIPS and $\Delta\beta\text{-}{\rm IPS}$ significantly improving MSE, C.I. width and statistical power over $\Delta$-IPS.

\subsection{Learning improved policies (RQ3)}
We follow a similar setup as described in earlier work~\cite{Jeunen2024_MultiObjective}, to learn a policy with a continuous multivariate action space corresponding to scalarisation weights in a multi-objective recommendation setting on a large-scale short-video platform, with over 160 million monthly active users.
We learn policies that maximise a lower bound on policy value for pointwise and pairwise estimators, and deploy them in two-week A/B tests, measuring key metrics to the platform.
Policies aim to maximise a learnt reward metric that aims to maximise statistical power w.r.t. the North Star~\cite{Jeunen2024_Learning}.
Results for these A/B-tests are reported in Table~\ref{tab:online_experiment}.
Whilst both methods are effective at improving the target metric---the policy optimising the $\Delta$-OPE lower bound increases the treatment effect estimate, and moves more metrics.
These results highlight the promise of our proposed methods.

\section{Conclusions \& Outlook}
This work has introduced the $\Delta$-OPE task, advocating for \emph{pairwise} policy value comparisons instead of \emph{pointwise} estimates.
We motivate this from the insight that the \emph{difference} in policy values can often be estimated with tighter confidence intervals if the policies have positive covariance.
We introduce $\Delta$-OPE formulations for common estimators, deriving the optimal variance-minimising global additive control variate for the IPS family.

Through experiments leveraging reproducible simulation environments as well as real-world end-users, we verify the effectiveness of our proposed methods in both evaluation and learning tasks.
All source code to reproduce the synthetic experiments can be found at \href{https://github.com/olivierjeunen/delta-OPE-recsys-2024}{github.com/olivierjeunen/delta-OPE-recsys-2024}.
Future work can expand the $\Delta$-OPE view towards doubly robust estimators~\cite{Dudik2014,Farajtabar2018,su2020doubly} and ranking applications~\cite{Gupta2024}.

\bibliographystyle{ACM-Reference-Format}
\bibliography{bibliography}


\begin{thebibliography}{52}


\ifx \showCODEN    \undefined \def \showCODEN     #1{\unskip}     \fi
\ifx \showDOI      \undefined \def \showDOI       #1{#1}\fi
\ifx \showISBNx    \undefined \def \showISBNx     #1{\unskip}     \fi
\ifx \showISBNxiii \undefined \def \showISBNxiii  #1{\unskip}     \fi
\ifx \showISSN     \undefined \def \showISSN      #1{\unskip}     \fi
\ifx \showLCCN     \undefined \def \showLCCN      #1{\unskip}     \fi
\ifx \shownote     \undefined \def \shownote      #1{#1}          \fi
\ifx \showarticletitle \undefined \def \showarticletitle #1{#1}   \fi
\ifx \showURL      \undefined \def \showURL       {\relax}        \fi
\providecommand\bibfield[2]{#2}
\providecommand\bibinfo[2]{#2}
\providecommand\natexlab[1]{#1}
\providecommand\showeprint[2][]{arXiv:#2}

\bibitem[Bottou et~al\mbox{.}(2013)]%
        {Bottou2013}
\bibfield{author}{\bibinfo{person}{L{{\'e}}on Bottou}, \bibinfo{person}{Jonas Peters}, \bibinfo{person}{Joaquin Qui{{\~n}}onero-Candela}, \bibinfo{person}{Denis~X. Charles}, \bibinfo{person}{D.~Max Chickering}, \bibinfo{person}{Elon Portugaly}, \bibinfo{person}{Dipankar Ray}, \bibinfo{person}{Patrice Simard}, {and} \bibinfo{person}{Ed Snelson}.} \bibinfo{year}{2013}\natexlab{}.
\newblock \showarticletitle{Counterfactual Reasoning and Learning Systems: The Example of Computational Advertising}.
\newblock \bibinfo{journal}{\emph{Journal of Machine Learning Research}} \bibinfo{volume}{14}, \bibinfo{number}{101} (\bibinfo{year}{2013}), \bibinfo{pages}{3207--3260}.
\newblock
\urldef\tempurl%
\url{http://jmlr.org/papers/v14/bottou13a.html}
\showURL{%
\tempurl}


\bibitem[Chen et~al\mbox{.}(2019)]%
        {chen2019top}
\bibfield{author}{\bibinfo{person}{Minmin Chen}, \bibinfo{person}{Alex Beutel}, \bibinfo{person}{Paul Covington}, \bibinfo{person}{Sagar Jain}, \bibinfo{person}{Francois Belletti}, {and} \bibinfo{person}{Ed~H. Chi}.} \bibinfo{year}{2019}\natexlab{}.
\newblock \showarticletitle{Top-K Off-Policy Correction for a REINFORCE Recommender System}. In \bibinfo{booktitle}{\emph{Proceedings of the Twelfth ACM International Conference on Web Search and Data Mining}} \emph{(\bibinfo{series}{WSDM '19})}. \bibinfo{publisher}{ACM}, \bibinfo{pages}{456–464}.
\newblock
\showISBNx{9781450359405}
\urldef\tempurl%
\url{https://doi.org/10.1145/3289600.3290999}
\showDOI{\tempurl}


\bibitem[Chen et~al\mbox{.}(2021)]%
        {Chen2021}
\bibfield{author}{\bibinfo{person}{Minmin Chen}, \bibinfo{person}{Bo Chang}, \bibinfo{person}{Can Xu}, {and} \bibinfo{person}{Ed~H. Chi}.} \bibinfo{year}{2021}\natexlab{}.
\newblock \showarticletitle{User Response Models to Improve a REINFORCE Recommender System}. In \bibinfo{booktitle}{\emph{Proceedings of the 14th ACM International Conference on Web Search and Data Mining}} \emph{(\bibinfo{series}{WSDM '21})}. \bibinfo{publisher}{ACM}, \bibinfo{pages}{121–129}.
\newblock
\showISBNx{9781450382977}
\urldef\tempurl%
\url{https://doi.org/10.1145/3437963.3441764}
\showDOI{\tempurl}


\bibitem[Chen et~al\mbox{.}(2022)]%
        {chen2022actorcritic}
\bibfield{author}{\bibinfo{person}{Minmin Chen}, \bibinfo{person}{Can Xu}, \bibinfo{person}{Vince Gatto}, \bibinfo{person}{Devanshu Jain}, \bibinfo{person}{Aviral Kumar}, {and} \bibinfo{person}{Ed Chi}.} \bibinfo{year}{2022}\natexlab{}.
\newblock \showarticletitle{Off-Policy Actor-Critic for Recommender Systems}. In \bibinfo{booktitle}{\emph{Proceedings of the 16th ACM Conference on Recommender Systems}} \emph{(\bibinfo{series}{RecSys '22})}. \bibinfo{publisher}{ACM}, \bibinfo{pages}{338–349}.
\newblock
\showISBNx{9781450392785}
\urldef\tempurl%
\url{https://doi.org/10.1145/3523227.3546758}
\showDOI{\tempurl}


\bibitem[Dayan(1991)]%
        {Dayan1991}
\bibfield{author}{\bibinfo{person}{Peter Dayan}.} \bibinfo{year}{1991}\natexlab{}.
\newblock \showarticletitle{Reinforcement Comparison}.
\newblock In \bibinfo{booktitle}{\emph{Connectionist Models}}, \bibfield{editor}{\bibinfo{person}{David~S. Touretzky}, \bibinfo{person}{Jeffrey~L. Elman}, \bibinfo{person}{Terrence~J. Sejnowski}, {and} \bibinfo{person}{Geoffrey~E. Hinton}} (Eds.). \bibinfo{publisher}{Morgan Kaufmann}, \bibinfo{pages}{45--51}.
\newblock
\showISBNx{978-1-4832-1448-1}
\urldef\tempurl%
\url{https://doi.org/10.1016/B978-1-4832-1448-1.50011-1}
\showDOI{\tempurl}


\bibitem[Dong et~al\mbox{.}(2020)]%
        {Dong2020}
\bibfield{author}{\bibinfo{person}{Zhenhua Dong}, \bibinfo{person}{Hong Zhu}, \bibinfo{person}{Pengxiang Cheng}, \bibinfo{person}{Xinhua Feng}, \bibinfo{person}{Guohao Cai}, \bibinfo{person}{Xiuqiang He}, \bibinfo{person}{Jun Xu}, {and} \bibinfo{person}{Jirong Wen}.} \bibinfo{year}{2020}\natexlab{}.
\newblock \showarticletitle{Counterfactual learning for recommender system}. In \bibinfo{booktitle}{\emph{Proceedings of the 14th ACM Conference on Recommender Systems}} \emph{(\bibinfo{series}{RecSys '20})}. \bibinfo{publisher}{ACM}, \bibinfo{pages}{568–569}.
\newblock
\showISBNx{9781450375832}
\urldef\tempurl%
\url{https://doi.org/10.1145/3383313.3411552}
\showDOI{\tempurl}


\bibitem[Dud{\'i}k et~al\mbox{.}(2014)]%
        {Dudik2014}
\bibfield{author}{\bibinfo{person}{Miroslav Dud{\'i}k}, \bibinfo{person}{Dumitru Erhan}, \bibinfo{person}{John Langford}, {and} \bibinfo{person}{Lihong Li}.} \bibinfo{year}{2014}\natexlab{}.
\newblock \showarticletitle{{Doubly Robust Policy Evaluation and Optimization}}.
\newblock \bibinfo{journal}{\emph{Statist. Sci.}} \bibinfo{volume}{29}, \bibinfo{number}{4} (\bibinfo{year}{2014}), \bibinfo{pages}{485 -- 511}.
\newblock
\urldef\tempurl%
\url{https://doi.org/10.1214/14-STS500}
\showDOI{\tempurl}


\bibitem[Farajtabar et~al\mbox{.}(2018)]%
        {Farajtabar2018}
\bibfield{author}{\bibinfo{person}{Mehrdad Farajtabar}, \bibinfo{person}{Yinlam Chow}, {and} \bibinfo{person}{Mohammad Ghavamzadeh}.} \bibinfo{year}{2018}\natexlab{}.
\newblock \showarticletitle{More Robust Doubly Robust Off-policy Evaluation}. In \bibinfo{booktitle}{\emph{Proceedings of the 35th International Conference on Machine Learning}} \emph{(\bibinfo{series}{Proceedings of Machine Learning Research}, Vol.~\bibinfo{volume}{80})}, \bibfield{editor}{\bibinfo{person}{Jennifer Dy} {and} \bibinfo{person}{Andreas Krause}} (Eds.). \bibinfo{publisher}{PMLR}, \bibinfo{pages}{1447--1456}.
\newblock
\urldef\tempurl%
\url{https://proceedings.mlr.press/v80/farajtabar18a.html}
\showURL{%
\tempurl}


\bibitem[Faury et~al\mbox{.}(2020)]%
        {Faury2020}
\bibfield{author}{\bibinfo{person}{Louis Faury}, \bibinfo{person}{Ugo Tanielian}, \bibinfo{person}{Elvis Dohmatob}, \bibinfo{person}{Elena Smirnova}, {and} \bibinfo{person}{Flavian Vasile}.} \bibinfo{year}{2020}\natexlab{}.
\newblock \showarticletitle{Distributionally Robust Counterfactual Risk Minimization}.
\newblock \bibinfo{journal}{\emph{Proceedings of the AAAI Conference on Artificial Intelligence}} \bibinfo{volume}{34}, \bibinfo{number}{04} (\bibinfo{date}{Apr.} \bibinfo{year}{2020}), \bibinfo{pages}{3850--3857}.
\newblock
\urldef\tempurl%
\url{https://doi.org/10.1609/aaai.v34i04.5797}
\showDOI{\tempurl}


\bibitem[Gilotte et~al\mbox{.}(2018)]%
        {Gilotte2018}
\bibfield{author}{\bibinfo{person}{Alexandre Gilotte}, \bibinfo{person}{Cl\'{e}ment Calauz\`{e}nes}, \bibinfo{person}{Thomas Nedelec}, \bibinfo{person}{Alexandre Abraham}, {and} \bibinfo{person}{Simon Doll\'{e}}.} \bibinfo{year}{2018}\natexlab{}.
\newblock \showarticletitle{Offline A/B Testing for Recommender Systems}. In \bibinfo{booktitle}{\emph{Proc. of the Eleventh ACM International Conference on Web Search and Data Mining}} \emph{(\bibinfo{series}{WSDM '18})}. \bibinfo{publisher}{ACM}, \bibinfo{pages}{198–206}.
\newblock
\urldef\tempurl%
\url{https://doi.org/10.1145/3159652.3159687}
\showDOI{\tempurl}


\bibitem[Greensmith et~al\mbox{.}(2004)]%
        {Greensmith2004}
\bibfield{author}{\bibinfo{person}{Evan Greensmith}, \bibinfo{person}{Peter~L. Bartlett}, {and} \bibinfo{person}{Jonathan Baxter}.} \bibinfo{year}{2004}\natexlab{}.
\newblock \showarticletitle{Variance Reduction Techniques for Gradient Estimates in Reinforcement Learning}.
\newblock \bibinfo{journal}{\emph{J. Mach. Learn. Res.}}  \bibinfo{volume}{5} (\bibinfo{date}{dec} \bibinfo{year}{2004}), \bibinfo{pages}{1471–1530}.
\newblock
\showISSN{1532-4435}


\bibitem[Gruson et~al\mbox{.}(2019)]%
        {Gruson2019}
\bibfield{author}{\bibinfo{person}{Alois Gruson}, \bibinfo{person}{Praveen Chandar}, \bibinfo{person}{Christophe Charbuillet}, \bibinfo{person}{James McInerney}, \bibinfo{person}{Samantha Hansen}, \bibinfo{person}{Damien Tardieu}, {and} \bibinfo{person}{Ben Carterette}.} \bibinfo{year}{2019}\natexlab{}.
\newblock \showarticletitle{Offline Evaluation to Make Decisions About PlaylistRecommendation Algorithms}. In \bibinfo{booktitle}{\emph{Proceedings of the Twelfth ACM International Conference on Web Search and Data Mining}} \emph{(\bibinfo{series}{WSDM '19})}. \bibinfo{publisher}{ACM}, \bibinfo{pages}{420–428}.
\newblock
\showISBNx{9781450359405}
\urldef\tempurl%
\url{https://doi.org/10.1145/3289600.3291027}
\showDOI{\tempurl}


\bibitem[Gupta et~al\mbox{.}(2024a)]%
        {Gupta2024}
\bibfield{author}{\bibinfo{person}{Shashank Gupta}, \bibinfo{person}{Philipp Hager}, \bibinfo{person}{Jin Huang}, \bibinfo{person}{Ali Vardasbi}, {and} \bibinfo{person}{Harrie Oosterhuis}.} \bibinfo{year}{2024}\natexlab{a}.
\newblock \showarticletitle{Unbiased Learning to Rank: On Recent Advances and Practical Applications}. In \bibinfo{booktitle}{\emph{Proceedings of the 17th ACM International Conference on Web Search and Data Mining}} \emph{(\bibinfo{series}{WSDM '24})}. \bibinfo{publisher}{ACM}, \bibinfo{pages}{1118–1121}.
\newblock
\showISBNx{9798400703713}
\urldef\tempurl%
\url{https://doi.org/10.1145/3616855.3636451}
\showDOI{\tempurl}


\bibitem[Gupta et~al\mbox{.}(2024b)]%
        {Gupta2024_OptimalBaselines}
\bibfield{author}{\bibinfo{person}{Shashank Gupta}, \bibinfo{person}{Olivier Jeunen}, \bibinfo{person}{Harrie Oosterhuis}, {and} \bibinfo{person}{Maarten de Rijke}.} \bibinfo{year}{2024}\natexlab{b}.
\newblock \showarticletitle{Optimal Baseline Corrections for Off-Policy Contextual Bandits}. In \bibinfo{booktitle}{\emph{Proceedings of the 18th ACM Conference on Recommender Systems}} \emph{(\bibinfo{series}{RecSys '24})}.
\newblock
\showeprint[arxiv]{2405.05736}~[cs.LG]


\bibitem[Gupta et~al\mbox{.}(2023a)]%
        {gupta2023deep}
\bibfield{author}{\bibinfo{person}{Shashank Gupta}, \bibinfo{person}{Harrie Oosterhuis}, {and} \bibinfo{person}{Maarten de Rijke}.} \bibinfo{year}{2023}\natexlab{a}.
\newblock \showarticletitle{A Deep Generative Recommendation Method for Unbiased Learning from Implicit Feedback}. In \bibinfo{booktitle}{\emph{Proceedings of the 2023 ACM SIGIR International Conference on Theory of Information Retrieval}}. \bibinfo{pages}{87--93}.
\newblock


\bibitem[Gupta et~al\mbox{.}(2023b)]%
        {gupta2023safe}
\bibfield{author}{\bibinfo{person}{Shashank Gupta}, \bibinfo{person}{Harrie Oosterhuis}, {and} \bibinfo{person}{Maarten de Rijke}.} \bibinfo{year}{2023}\natexlab{b}.
\newblock \showarticletitle{Safe deployment for counterfactual learning to rank with exposure-based risk minimization}. In \bibinfo{booktitle}{\emph{Proceedings of the 46th International ACM SIGIR Conference on Research and Development in Information Retrieval}}. \bibinfo{pages}{249--258}.
\newblock


\bibitem[Horvitz and Thompson(1952)]%
        {Horvitz1952}
\bibfield{author}{\bibinfo{person}{Daniel~G. Horvitz} {and} \bibinfo{person}{Donovan~J. Thompson}.} \bibinfo{year}{1952}\natexlab{}.
\newblock \showarticletitle{A Generalization of Sampling Without Replacement from a Finite Universe}.
\newblock \bibinfo{journal}{\emph{J. Amer. Statist. Assoc.}} \bibinfo{volume}{47}, \bibinfo{number}{260} (\bibinfo{year}{1952}), \bibinfo{pages}{663--685}.
\newblock
\urldef\tempurl%
\url{https://doi.org/10.1080/01621459.1952.10483446}
\showDOI{\tempurl}


\bibitem[Jeunen(2021)]%
        {Jeunen2021_Thesis}
\bibfield{author}{\bibinfo{person}{Olivier Jeunen}.} \bibinfo{year}{2021}\natexlab{}.
\newblock \emph{\bibinfo{title}{Offline approaches to recommendation with online success}}.
\newblock \bibinfo{thesistype}{Ph.\,D. Dissertation}. \bibinfo{school}{University of Antwerp}.
\newblock


\bibitem[Jeunen and Goethals(2020)]%
        {Jeunen2020REVEAL}
\bibfield{author}{\bibinfo{person}{Olivier Jeunen} {and} \bibinfo{person}{Bart Goethals}.} \bibinfo{year}{2020}\natexlab{}.
\newblock \showarticletitle{An Empirical Evaluation of Doubly Robust Learning for Recommendation}. In \bibinfo{booktitle}{\emph{Proc. of the ACM RecSys Workshop on Bandit Learning from User Interactions}} \emph{(\bibinfo{series}{REVEAL '20})}.
\newblock


\bibitem[Jeunen and Goethals(2021)]%
        {Jeunen2021_Pessimism}
\bibfield{author}{\bibinfo{person}{Olivier Jeunen} {and} \bibinfo{person}{Bart Goethals}.} \bibinfo{year}{2021}\natexlab{}.
\newblock \showarticletitle{Pessimistic Reward Models for Off-Policy Learning in Recommendation}. In \bibinfo{booktitle}{\emph{Proceedings of the 15th ACM Conference on Recommender Systems}} \emph{(\bibinfo{series}{RecSys '21})}. \bibinfo{publisher}{ACM}, \bibinfo{pages}{63–74}.
\newblock
\showISBNx{9781450384582}
\urldef\tempurl%
\url{https://doi.org/10.1145/3460231.3474247}
\showDOI{\tempurl}


\bibitem[Jeunen and Goethals(2023)]%
        {Jeunen2023}
\bibfield{author}{\bibinfo{person}{Olivier Jeunen} {and} \bibinfo{person}{Bart Goethals}.} \bibinfo{year}{2023}\natexlab{}.
\newblock \showarticletitle{Pessimistic Decision-Making for Recommender Systems}.
\newblock \bibinfo{journal}{\emph{ACM Trans. Recomm. Syst.}} \bibinfo{volume}{1}, \bibinfo{number}{1}, Article \bibinfo{articleno}{4} (\bibinfo{date}{feb} \bibinfo{year}{2023}), \bibinfo{numpages}{27}~pages.
\newblock
\urldef\tempurl%
\url{https://doi.org/10.1145/3568029}
\showDOI{\tempurl}


\bibitem[Jeunen et~al\mbox{.}(2022)]%
        {CONSEQUENCES2022}
\bibfield{author}{\bibinfo{person}{Olivier Jeunen}, \bibinfo{person}{Thorsten Joachims}, \bibinfo{person}{Harrie Oosterhuis}, \bibinfo{person}{Yuta Saito}, {and} \bibinfo{person}{Flavian Vasile}.} \bibinfo{year}{2022}\natexlab{}.
\newblock \showarticletitle{CONSEQUENCES — Causality, Counterfactuals and Sequential Decision-Making for Recommender Systems}. In \bibinfo{booktitle}{\emph{Proceedings of the 16th ACM Conference on Recommender Systems}} \emph{(\bibinfo{series}{RecSys '22})}. \bibinfo{publisher}{ACM}, \bibinfo{pages}{654–657}.
\newblock
\showISBNx{9781450392785}
\urldef\tempurl%
\url{https://doi.org/10.1145/3523227.3547409}
\showDOI{\tempurl}


\bibitem[Jeunen and London(2023)]%
        {JeunenLondon2023}
\bibfield{author}{\bibinfo{person}{Olivier Jeunen} {and} \bibinfo{person}{Ben London}.} \bibinfo{year}{2023}\natexlab{}.
\newblock \showarticletitle{Offline Recommender System Evaluation under Unobserved Confounding}. In \bibinfo{booktitle}{\emph{RecSys workshop on Causality, Counterfactuals and Sequential Decision-Making}} \emph{(\bibinfo{series}{CONSEQUENCES '24})}.
\newblock
\showeprint[arxiv]{2309.04222}~[cs.LG]


\bibitem[Jeunen et~al\mbox{.}(2024)]%
        {Jeunen2024_MultiObjective}
\bibfield{author}{\bibinfo{person}{Olivier Jeunen}, \bibinfo{person}{Jatin Mandav}, \bibinfo{person}{Ivan Potapov}, \bibinfo{person}{Nakul Agarwal}, \bibinfo{person}{Sourabh Vaid}, \bibinfo{person}{Wenzhe Shi}, {and} \bibinfo{person}{Aleksei Ustimenko}.} \bibinfo{year}{2024}\natexlab{}.
\newblock \bibinfo{title}{Multi-Objective Recommendation via Multivariate Policy Learning}.
\newblock
\newblock
\showeprint[arxiv]{2405.02141}~[cs.IR]


\bibitem[Jeunen et~al\mbox{.}(2020)]%
        {Jeunen2020}
\bibfield{author}{\bibinfo{person}{Olivier Jeunen}, \bibinfo{person}{David Rohde}, \bibinfo{person}{Flavian Vasile}, {and} \bibinfo{person}{Martin Bompaire}.} \bibinfo{year}{2020}\natexlab{}.
\newblock \showarticletitle{Joint Policy-Value Learning for Recommendation}. In \bibinfo{booktitle}{\emph{Proceedings of the 26th ACM SIGKDD International Conference on Knowledge Discovery \& Data Mining}} \emph{(\bibinfo{series}{KDD '20})}. \bibinfo{publisher}{ACM}, \bibinfo{pages}{1223–1233}.
\newblock
\showISBNx{9781450379984}
\urldef\tempurl%
\url{https://doi.org/10.1145/3394486.3403175}
\showDOI{\tempurl}


\bibitem[Jeunen and Ustimenko(2024)]%
        {Jeunen2024_Learning}
\bibfield{author}{\bibinfo{person}{Olivier Jeunen} {and} \bibinfo{person}{Aleksei Ustimenko}.} \bibinfo{year}{2024}\natexlab{}.
\newblock \showarticletitle{Learning Metrics that Maximise Power for Accelerated A/B-Tests}. In \bibinfo{booktitle}{\emph{Proceedings of the 26th ACM SIGKDD International Conference on Knowledge Discovery \& Data Mining}} \emph{(\bibinfo{series}{KDD '24})}.
\newblock
\showeprint[arxiv]{2402.03915}~[cs.LG]


\bibitem[Joachims et~al\mbox{.}(2018a)]%
        {Joachims2018}
\bibfield{author}{\bibinfo{person}{Thorsten Joachims}, \bibinfo{person}{Adith Swaminathan}, {and} \bibinfo{person}{Maarten de Rijke}.} \bibinfo{year}{2018}\natexlab{a}.
\newblock \showarticletitle{Deep Learning with Logged Bandit Feedback}. In \bibinfo{booktitle}{\emph{International Conference on Learning Representations}}.
\newblock
\urldef\tempurl%
\url{https://openreview.net/forum?id=SJaP_-xAb}
\showURL{%
\tempurl}


\bibitem[Joachims et~al\mbox{.}(2018b)]%
        {Joachims2018_REVEAL}
\bibfield{author}{\bibinfo{person}{Thorsten Joachims}, \bibinfo{person}{Adith Swaminathan}, \bibinfo{person}{Yves Raimond}, \bibinfo{person}{Olivier Koch}, {and} \bibinfo{person}{Flavian Vasile}.} \bibinfo{year}{2018}\natexlab{b}.
\newblock \showarticletitle{REVEAL 2018: offline evaluation for recommender systems}. In \bibinfo{booktitle}{\emph{Proceedings of the 12th ACM Conference on Recommender Systems}} (Vancouver, British Columbia, Canada) \emph{(\bibinfo{series}{RecSys '18})}. \bibinfo{publisher}{Association for Computing Machinery}, \bibinfo{address}{New York, NY, USA}, \bibinfo{pages}{514–515}.
\newblock
\showISBNx{9781450359016}
\urldef\tempurl%
\url{https://doi.org/10.1145/3240323.3240334}
\showDOI{\tempurl}


\bibitem[Kong(1992)]%
        {Kong1992}
\bibfield{author}{\bibinfo{person}{Augustine Kong}.} \bibinfo{year}{1992}\natexlab{}.
\newblock \showarticletitle{A note on importance sampling using standardized weights}.
\newblock \bibinfo{journal}{\emph{University of Chicago, Dept. of Statistics, Tech. Rep}}  \bibinfo{volume}{348} (\bibinfo{year}{1992}).
\newblock


\bibitem[Lehmann and Romano(2005)]%
        {Lehmann2005}
\bibfield{author}{\bibinfo{person}{Erich~L Lehmann} {and} \bibinfo{person}{Joseph~P Romano}.} \bibinfo{year}{2005}\natexlab{}.
\newblock \bibinfo{title}{Testing statistical hypotheses}.
\newblock
\newblock


\bibitem[Li et~al\mbox{.}(2010)]%
        {Li2010}
\bibfield{author}{\bibinfo{person}{Lihong Li}, \bibinfo{person}{Wei Chu}, \bibinfo{person}{John Langford}, {and} \bibinfo{person}{Robert~E. Schapire}.} \bibinfo{year}{2010}\natexlab{}.
\newblock \showarticletitle{A contextual-bandit approach to personalized news article recommendation}. In \bibinfo{booktitle}{\emph{Proceedings of the 19th International Conference on World Wide Web}} \emph{(\bibinfo{series}{WWW '10})}. \bibinfo{publisher}{ACM}, \bibinfo{pages}{661–670}.
\newblock
\showISBNx{9781605587998}
\urldef\tempurl%
\url{https://doi.org/10.1145/1772690.1772758}
\showDOI{\tempurl}


\bibitem[Liu et~al\mbox{.}(2020)]%
        {Liu2020}
\bibfield{author}{\bibinfo{person}{Dugang Liu}, \bibinfo{person}{Pengxiang Cheng}, \bibinfo{person}{Zhenhua Dong}, \bibinfo{person}{Xiuqiang He}, \bibinfo{person}{Weike Pan}, {and} \bibinfo{person}{Zhong Ming}.} \bibinfo{year}{2020}\natexlab{}.
\newblock \showarticletitle{A General Knowledge Distillation Framework for Counterfactual Recommendation via Uniform Data}. In \bibinfo{booktitle}{\emph{Proceedings of the 43rd International ACM SIGIR Conference on Research and Development in Information Retrieval}} \emph{(\bibinfo{series}{SIGIR '20})}. \bibinfo{publisher}{ACM}, \bibinfo{pages}{831–840}.
\newblock
\showISBNx{9781450380164}
\urldef\tempurl%
\url{https://doi.org/10.1145/3397271.3401083}
\showDOI{\tempurl}


\bibitem[Liu et~al\mbox{.}(2022)]%
        {Liu2022}
\bibfield{author}{\bibinfo{person}{Yaxu Liu}, \bibinfo{person}{Jui-Nan Yen}, \bibinfo{person}{Bowen Yuan}, \bibinfo{person}{Rundong Shi}, \bibinfo{person}{Peng Yan}, {and} \bibinfo{person}{Chih-Jen Lin}.} \bibinfo{year}{2022}\natexlab{}.
\newblock \showarticletitle{Practical Counterfactual Policy Learning for Top-K Recommendations}. In \bibinfo{booktitle}{\emph{Proceedings of the 28th ACM SIGKDD Conference on Knowledge Discovery and Data Mining}} \emph{(\bibinfo{series}{KDD '22})}. \bibinfo{publisher}{ACM}, \bibinfo{pages}{1141–1151}.
\newblock
\showISBNx{9781450393850}
\urldef\tempurl%
\url{https://doi.org/10.1145/3534678.3539295}
\showDOI{\tempurl}


\bibitem[Ma et~al\mbox{.}(2020)]%
        {ma2020off}
\bibfield{author}{\bibinfo{person}{Jiaqi Ma}, \bibinfo{person}{Zhe Zhao}, \bibinfo{person}{Xinyang Yi}, \bibinfo{person}{Ji Yang}, \bibinfo{person}{Minmin Chen}, \bibinfo{person}{Jiaxi Tang}, \bibinfo{person}{Lichan Hong}, {and} \bibinfo{person}{Ed~H. Chi}.} \bibinfo{year}{2020}\natexlab{}.
\newblock \showarticletitle{Off-Policy Learning in Two-Stage Recommender Systems}. In \bibinfo{booktitle}{\emph{Proceedings of The Web Conference 2020}} \emph{(\bibinfo{series}{WWW '20})}. \bibinfo{publisher}{ACM}, \bibinfo{pages}{463–473}.
\newblock
\showISBNx{9781450370233}
\urldef\tempurl%
\url{https://doi.org/10.1145/3366423.3380130}
\showDOI{\tempurl}


\bibitem[Mairesse et~al\mbox{.}(2021)]%
        {Mairesse2021}
\bibfield{author}{\bibinfo{person}{Francois Mairesse}, \bibinfo{person}{Zhonghao Luo}, {and} \bibinfo{person}{Tao Ye}.} \bibinfo{year}{2021}\natexlab{}.
\newblock \showarticletitle{Learning a Voice-based Conversational Recommender using Offline Policy Optimization}. In \bibinfo{booktitle}{\emph{Proceedings of the 15th ACM Conference on Recommender Systems}} \emph{(\bibinfo{series}{RecSys '21})}. \bibinfo{publisher}{ACM}, \bibinfo{pages}{562–564}.
\newblock
\showISBNx{9781450384582}
\urldef\tempurl%
\url{https://doi.org/10.1145/3460231.3474600}
\showDOI{\tempurl}


\bibitem[Maurer and Pontil(2009)]%
        {Maurer2009}
\bibfield{author}{\bibinfo{person}{Andreas Maurer} {and} \bibinfo{person}{Massimiliano Pontil}.} \bibinfo{year}{2009}\natexlab{}.
\newblock \showarticletitle{Empirical Bernstein Bounds and Sample Variance Penalization}.
\newblock \bibinfo{journal}{\emph{Stat.}}  \bibinfo{volume}{1050} (\bibinfo{year}{2009}), \bibinfo{pages}{21}.
\newblock


\bibitem[Owen(2013)]%
        {Owen2013}
\bibfield{author}{\bibinfo{person}{Art~B. Owen}.} \bibinfo{year}{2013}\natexlab{}.
\newblock \bibinfo{booktitle}{\emph{Monte Carlo theory, methods and examples}}.
\newblock


\bibitem[Sagtani et~al\mbox{.}(2024)]%
        {Sagtani2024}
\bibfield{author}{\bibinfo{person}{Hitesh Sagtani}, \bibinfo{person}{Madan~Gopal Jhawar}, \bibinfo{person}{Rishabh Mehrotra}, {and} \bibinfo{person}{Olivier Jeunen}.} \bibinfo{year}{2024}\natexlab{}.
\newblock \showarticletitle{Ad-load Balancing via Off-policy Learning in a Content Marketplace}. In \bibinfo{booktitle}{\emph{Proceedings of the 17th ACM International Conference on Web Search and Data Mining}} \emph{(\bibinfo{series}{WSDM '24})}. \bibinfo{publisher}{ACM}, \bibinfo{pages}{586–595}.
\newblock
\showISBNx{9798400703713}
\urldef\tempurl%
\url{https://doi.org/10.1145/3616855.3635846}
\showDOI{\tempurl}


\bibitem[Saito et~al\mbox{.}(2021)]%
        {Saito2021_OBP}
\bibfield{author}{\bibinfo{person}{Yuta Saito}, \bibinfo{person}{Shunsuke Aihara}, \bibinfo{person}{Megumi Matsutani}, {and} \bibinfo{person}{Yusuke Narita}.} \bibinfo{year}{2021}\natexlab{}.
\newblock \showarticletitle{Open Bandit Dataset and Pipeline: Towards Realistic and Reproducible Off-Policy Evaluation}. In \bibinfo{booktitle}{\emph{Proceedings of the Neural Information Processing Systems Track on Datasets and Benchmarks}}, Vol.~\bibinfo{volume}{1}.
\newblock


\bibitem[Saito and Joachims(2021)]%
        {Saito2021}
\bibfield{author}{\bibinfo{person}{Yuta Saito} {and} \bibinfo{person}{Thorsten Joachims}.} \bibinfo{year}{2021}\natexlab{}.
\newblock \showarticletitle{Counterfactual Learning and Evaluation for Recommender Systems: Foundations, Implementations, and Recent Advances}. In \bibinfo{booktitle}{\emph{Proc. of the 15th ACM Conference on Recommender Systems}} \emph{(\bibinfo{series}{RecSys '21})}. \bibinfo{publisher}{ACM}, \bibinfo{pages}{828–830}.
\newblock
\showISBNx{9781450384582}
\urldef\tempurl%
\url{https://doi.org/10.1145/3460231.3473320}
\showDOI{\tempurl}


\bibitem[Schulman et~al\mbox{.}(2015)]%
        {Schulman2015}
\bibfield{author}{\bibinfo{person}{John Schulman}, \bibinfo{person}{Sergey Levine}, \bibinfo{person}{Pieter Abbeel}, \bibinfo{person}{Michael Jordan}, {and} \bibinfo{person}{Philipp Moritz}.} \bibinfo{year}{2015}\natexlab{}.
\newblock \showarticletitle{Trust Region Policy Optimization}. In \bibinfo{booktitle}{\emph{Proceedings of the 32nd International Conference on Machine Learning}} \emph{(\bibinfo{series}{Proceedings of Machine Learning Research}, Vol.~\bibinfo{volume}{37})}, \bibfield{editor}{\bibinfo{person}{Francis Bach} {and} \bibinfo{person}{David Blei}} (Eds.). \bibinfo{publisher}{PMLR}, \bibinfo{address}{Lille, France}, \bibinfo{pages}{1889--1897}.
\newblock
\urldef\tempurl%
\url{https://proceedings.mlr.press/v37/schulman15.html}
\showURL{%
\tempurl}


\bibitem[Schulman et~al\mbox{.}(2017)]%
        {Schulman2017}
\bibfield{author}{\bibinfo{person}{John Schulman}, \bibinfo{person}{Filip Wolski}, \bibinfo{person}{Prafulla Dhariwal}, \bibinfo{person}{Alec Radford}, {and} \bibinfo{person}{Oleg Klimov}.} \bibinfo{year}{2017}\natexlab{}.
\newblock \bibinfo{title}{Proximal Policy Optimization Algorithms}.
\newblock
\newblock
\showeprint[arxiv]{1707.06347}~[cs.LG]


\bibitem[Si et~al\mbox{.}(2020)]%
        {Si2020}
\bibfield{author}{\bibinfo{person}{Nian Si}, \bibinfo{person}{Fan Zhang}, \bibinfo{person}{Zhengyuan Zhou}, {and} \bibinfo{person}{Jose Blanchet}.} \bibinfo{year}{2020}\natexlab{}.
\newblock \showarticletitle{Distributionally Robust Policy Evaluation and Learning in Offline Contextual Bandits}. In \bibinfo{booktitle}{\emph{Proceedings of the 37th International Conference on Machine Learning}} \emph{(\bibinfo{series}{Proceedings of Machine Learning Research}, Vol.~\bibinfo{volume}{119})}, \bibfield{editor}{\bibinfo{person}{Hal~Daumé III} {and} \bibinfo{person}{Aarti Singh}} (Eds.). \bibinfo{publisher}{PMLR}, \bibinfo{pages}{8884--8894}.
\newblock
\urldef\tempurl%
\url{https://proceedings.mlr.press/v119/si20a.html}
\showURL{%
\tempurl}


\bibitem[Su et~al\mbox{.}(2020)]%
        {su2020doubly}
\bibfield{author}{\bibinfo{person}{Yi Su}, \bibinfo{person}{Maria Dimakopoulou}, \bibinfo{person}{Akshay Krishnamurthy}, {and} \bibinfo{person}{Miroslav Dud{\'\i}k}.} \bibinfo{year}{2020}\natexlab{}.
\newblock \showarticletitle{Doubly robust off-policy evaluation with shrinkage}. In \bibinfo{booktitle}{\emph{International Conference on Machine Learning}}. PMLR, \bibinfo{pages}{9167--9176}.
\newblock


\bibitem[Swaminathan and Joachims(2015a)]%
        {Swaminathan2015_BLBF}
\bibfield{author}{\bibinfo{person}{Adith Swaminathan} {and} \bibinfo{person}{Thorsten Joachims}.} \bibinfo{year}{2015}\natexlab{a}.
\newblock \showarticletitle{Batch learning from logged bandit feedback through counterfactual risk minimization}.
\newblock \bibinfo{journal}{\emph{The Journal of Machine Learning Research}} \bibinfo{volume}{16}, \bibinfo{number}{1} (\bibinfo{year}{2015}), \bibinfo{pages}{1731--1755}.
\newblock


\bibitem[Swaminathan and Joachims(2015b)]%
        {Swaminathan2015}
\bibfield{author}{\bibinfo{person}{Adith Swaminathan} {and} \bibinfo{person}{Thorsten Joachims}.} \bibinfo{year}{2015}\natexlab{b}.
\newblock \showarticletitle{The Self-Normalized Estimator for Counterfactual Learning}. In \bibinfo{booktitle}{\emph{Advances in Neural Information Processing Systems}}, Vol.~\bibinfo{volume}{28}.
\newblock
\urldef\tempurl%
\url{https://proceedings.neurips.cc/paper_files/paper/2015/file/39027dfad5138c9ca0c474d71db915c3-Paper.pdf}
\showURL{%
\tempurl}


\bibitem[van~den Akker et~al\mbox{.}(2024)]%
        {vandenAkker2024}
\bibfield{author}{\bibinfo{person}{Bram van~den Akker}, \bibinfo{person}{Olivier Jeunen}, \bibinfo{person}{Ying Li}, \bibinfo{person}{Ben London}, \bibinfo{person}{Zahra Nazari}, {and} \bibinfo{person}{Devesh Parekh}.} \bibinfo{year}{2024}\natexlab{}.
\newblock \showarticletitle{Practical Bandits: An Industry Perspective}. In \bibinfo{booktitle}{\emph{Proceedings of the 17th ACM International Conference on Web Search and Data Mining}} \emph{(\bibinfo{series}{WSDM '24})}. \bibinfo{publisher}{ACM}, \bibinfo{pages}{1132–1135}.
\newblock
\showISBNx{9798400703713}
\urldef\tempurl%
\url{https://doi.org/10.1145/3616855.3636449}
\showDOI{\tempurl}


\bibitem[Vasile et~al\mbox{.}(2020)]%
        {Vasile2020}
\bibfield{author}{\bibinfo{person}{Flavian Vasile}, \bibinfo{person}{David Rohde}, \bibinfo{person}{Olivier Jeunen}, {and} \bibinfo{person}{Amine Benhalloum}.} \bibinfo{year}{2020}\natexlab{}.
\newblock \showarticletitle{A Gentle Introduction to Recommendation as Counterfactual Policy Learning}. In \bibinfo{booktitle}{\emph{Proceedings of the 28th ACM Conference on User Modeling, Adaptation and Personalization}} \emph{(\bibinfo{series}{UMAP '20})}. \bibinfo{publisher}{ACM}, \bibinfo{pages}{392–393}.
\newblock
\showISBNx{9781450368612}
\urldef\tempurl%
\url{https://doi.org/10.1145/3340631.3398666}
\showDOI{\tempurl}


\bibitem[Vasile et~al\mbox{.}(2021)]%
        {Vasile2021_DecisionTheory}
\bibfield{author}{\bibinfo{person}{Flavian Vasile}, \bibinfo{person}{David Rohde}, \bibinfo{person}{Olivier Jeunen}, \bibinfo{person}{Amine Benhalloum}, {and} \bibinfo{person}{Otmane Sakhi}.} \bibinfo{year}{2021}\natexlab{}.
\newblock \showarticletitle{Recommender Systems Through the Lens of Decision Theory}. In \bibinfo{booktitle}{\emph{Proceedings of the 30th World Wide Web Conference ACM Conference}}.
\newblock


\bibitem[Wan et~al\mbox{.}(2022)]%
        {wan2022safe}
\bibfield{author}{\bibinfo{person}{Runzhe Wan}, \bibinfo{person}{Branislav Kveton}, {and} \bibinfo{person}{Rui Song}.} \bibinfo{year}{2022}\natexlab{}.
\newblock \showarticletitle{Safe exploration for efficient policy evaluation and comparison}. In \bibinfo{booktitle}{\emph{International Conference on Machine Learning}}. PMLR, \bibinfo{pages}{22491--22511}.
\newblock


\bibitem[Williams(1988)]%
        {Williams1988}
\bibfield{author}{\bibinfo{person}{Ronald~J Williams}.} \bibinfo{year}{1988}\natexlab{}.
\newblock \showarticletitle{Toward a theory of reinforcement-learning connectionist systems}.
\newblock \bibinfo{journal}{\emph{Technical Report}} (\bibinfo{year}{1988}).
\newblock


\bibitem[Yang et~al\mbox{.}(2018)]%
        {Yang2018}
\bibfield{author}{\bibinfo{person}{Longqi Yang}, \bibinfo{person}{Yin Cui}, \bibinfo{person}{Yuan Xuan}, \bibinfo{person}{Chenyang Wang}, \bibinfo{person}{Serge Belongie}, {and} \bibinfo{person}{Deborah Estrin}.} \bibinfo{year}{2018}\natexlab{}.
\newblock \showarticletitle{Unbiased Offline Recommender Evaluation for Missing-Not-at-Random Implicit Feedback}. In \bibinfo{booktitle}{\emph{Proceedings of the 12th ACM Conference on Recommender Systems}} \emph{(\bibinfo{series}{RecSys '18})}. \bibinfo{publisher}{ACM}, \bibinfo{pages}{279–287}.
\newblock
\showISBNx{9781450359016}
\urldef\tempurl%
\url{https://doi.org/10.1145/3240323.3240355}
\showDOI{\tempurl}


\end{thebibliography}

\end{document}